\documentclass[10pt]{article}

\usepackage[preprint]{tmlr}    

\usepackage{amsmath,amssymb,amsthm}
\usepackage{booktabs}
\usepackage{graphicx}
\usepackage{xcolor}
\usepackage{caption}
\usepackage{hyperref}
\usepackage{url}
\graphicspath{{figures/}}

\newcommand{\tv}{\tau}
\newcommand{\base}{\theta_{0}}
\newcommand{\Real}{\mathbb{R}}
\newcommand{\E}{\mathbb{E}}
\newcommand{\indic}{\mathbf{1}}

\title{When Model Merging Rivals Joint Multi-Task Reinforcement Learning:
A Task-Vector Geometry Analysis\thanks{Code: \url{https://github.com/magicsquares137/maml-agent}
(branch \texttt{loop-fixes-and-throughput}); requires the AppWorld RL fork at
\url{https://github.com/magicsquares137/appworld-rl}.}}

\author{\name S.\ Aaron McClendon \email aaron.mcclendon@aimpointdigital.com \\
        \addr Aimpoint Digital Labs}

\begin{document}
\maketitle

\begin{abstract}
Model merging is promoted as a substitute for joint multi-task training, yet in the reinforcement-learning setting this substitution is essentially never tested against the baseline it claims to replace: methods merge independently released agents precisely because a joint model is unavailable. We build the missing comparison. Training difficulty-1 and difficulty-2 Qwen3-8B specialists on the AppWorld agent benchmark with LOOP, we merge them (TIES, RAM+) and pit the result against a jointly trained model on the same data. On task-goal completion, merging matches joint RL — and every merge variant is statistically indistinguishable. To explain why merge method does not matter here, we measure the geometry of the specialists' task vectors, which carries no task-sampling noise: they are near-orthogonal (cosine 0.06–0.10) despite ~65$\%$ support overlap, a small shared direction that grows over training and that we calibrate against a random-init floor and a same-run ceiling to confirm it reflects learning, not the low-rank parameterisation. Because direction and support are decoupled, support- and sign-based merging (RAM, TIES) collapse to near-uniform averaging. We release all code and statistics.
\end{abstract}

\section{Introduction}
\label{sec:intro}
Model merging promises to combine independently fine-tuned specialists into a single
multi-task model without joint retraining, by adding their \emph{task vectors}
(parameter offsets from a shared base) back onto the base
\citep{ilharco2023taskarithmetic}. A recent line of work extends merging to
\emph{reinforced} agentic models, where each specialist is trained with RL, and
proposes support- or subspace-aware merge rules such as RAM \citep{ram2026} and
Iso-CTS \citep{isocts2025}. A central appeal of the
merging setting is that it applies precisely when joint training is impossible --- when
the specialists come from different sources whose training data cannot be pooled.

That appeal has a corollary rarely tested: \emph{when data can be pooled, does merging
actually match joint multi-task RL?} We study this question in a controlled
within-benchmark setting on AppWorld \citep{trivedi2024appworld}, an interactive
digital-agent benchmark whose tasks are annotated by difficulty. We train two RL
specialists (difficulty-1 and difficulty-2), a jointly trained model on their union,
and two merges of the specialists, all from the same Qwen3-8B base with identical
recipe, and evaluate on the held-out \texttt{test\_normal} split.

Our findings are as follows:
\begin{enumerate}
\item \textbf{On the primary metric, everything ties.} On $168$ tasks, the specialists,
their merges, and the joint model are statistically indistinguishable in TGC. Only RL
over the base model is significant. At this scale ($8$B, $\sim\!10$ RL iterations,
$1$--$2$ task margins) the interesting comparisons live below the noise wall
(\S\ref{sec:results-perf}).
\item \textbf{Merging matches the baseline merging cannot run.} Merging is
indistinguishable from joint multi-task RL on the same data --- the comparison the
cross-domain setting forecloses (\S\ref{sec:results-perf}).
\item \textbf{A calibrated geometric diagnostic.} The specialists' task vectors are
near-orthogonal (cosine $0.06$--$0.10$) despite large support overlap, and this small
shared component \emph{develops} over training. We calibrate the measurement against a
floor and ceiling so it cannot be dismissed as a low-rank artifact, and use it to
explain why support- and sign-based merging have little to disentangle here
(\S\ref{sec:results-geom}, \S\ref{sec:results-calib}, \S\ref{sec:discussion}).
\item \textbf{A cautionary metric-sensitivity result.} A continuous partial-credit
metric resolves differences TGC cannot, and reports the merged models below the
un-merged ones. We show \emph{not} to over-read this: it reverses relative to TGC, it is
not emergent cross-task capability (joint does not exceed the per-task best specialist,
and a single specialist also beats the merges), and it cannot be closed by
trajectory-aware checkpoint selection (\S\ref{sec:results-metric}).
\end{enumerate}

We frame these findings as a calibrated null: a carefully bounded characterisation of
where merging suffices, a parameter-space mechanism that explains it, and a disciplined
account of the stronger claims we tested and set aside. Among these, we examine whether
selecting the merge checkpoint by an interference-aware objective could improve on naive
merging, a natural idea, and show analytically why it has no leverage in this regime,
which motivates the conditions under which it would (developed as future work,
\S\ref{sec:future}).

\section{Preliminaries}
\label{sec:prelim}

\subsection{Low-rank adaptation (LoRA)}
\label{sec:prelim-lora}
All models here are fine-tuned with LoRA \citep{hu2022lora}. For a frozen weight matrix
$W_0\in\Real^{d_{\text{out}}\times d_{\text{in}}}$, LoRA learns a low-rank update
\begin{equation}
W = W_0 + \Delta W,\qquad
\Delta W = \tfrac{\alpha}{r}\,B A,\quad
B\in\Real^{d_{\text{out}}\times r},\;A\in\Real^{r\times d_{\text{in}}},\;
r\ll\min(d_{\text{out}},d_{\text{in}}),
\end{equation}
with $A$ initialised randomly and $B$ initialised to $0$ (so $\Delta W=0$ at the start of
training), and $\alpha$ a fixed scale. Only $B,A$ are trained. We adapt $M$ target modules
(\S\ref{sec:setup-train}), each with its own factors $(B_i,A_i)$. After training we
\emph{merge the adapters into the base}, obtaining a full model $\theta$; the associated
\emph{task vector} $\tv=\theta-\base$ is zero on all unadapted parameters and, on adapted
module $i$, equals the low-rank update $\tfrac{\alpha}{r}B_iA_i$ ($i=1,\dots,M$). Two facts
about LoRA matter downstream. (i) Because every specialist adapts the \emph{same} $M$ target
modules, the supports of different task vectors (the coordinates they modify) coincide by
construction --- a point we control for. (ii) Each $\tfrac{\alpha}{r}B_iA_i$ is
rank-$\le r$; whether this constrains the \emph{direction} of $\tv$ relative to another
specialist's is an empirical question we calibrate in \S\ref{sec:results-geom}.

\subsection{PPO, RLOO, and LOOP}
\label{sec:prelim-rl}
Our specialists are trained with LOOP \citep{loop2025}, a critic-free policy-gradient
method for multi-turn LLM agents that combines a leave-one-out advantage with a PPO-style
clipped objective.

\paragraph{PPO.} Proximal Policy Optimisation \citep{schulman2017ppo} optimises a clipped
surrogate. With the importance ratio
$\rho_t(\theta)=\pi_\theta(a_t\mid s_t)/\pi_{\theta_{\text{old}}}(a_t\mid s_t)$ and an
advantage estimate $\hat A_t$,
\begin{equation}
\mathcal{L}^{\text{CLIP}}(\theta)=
\hat\E_t\!\Big[\min\big(\rho_t(\theta)\hat A_t,\;
\mathrm{clip}(\rho_t(\theta),1-\epsilon,1+\epsilon)\hat A_t\big)\Big].
\end{equation}

\paragraph{RLOO.} REINFORCE Leave-One-Out \citep{kool2019rloo,ahmadian2024rloo} removes
the value network by using the other samples as a baseline. For a prompt $x$, draw
$K$ trajectories $y_1,\dots,y_K\sim\pi_{\theta_{\text{old}}}(\cdot\mid x)$ with scalar
returns $R(y_i)$. The leave-one-out advantage is
\begin{equation}
\hat A_i \;=\; R(y_i)-\frac{1}{K-1}\sum_{j\neq i}R(y_j)
\;=\;\frac{K}{K-1}\Big(R(y_i)-\bar R\Big),\qquad
\bar R=\frac1K\sum_{k}R(y_k),
\end{equation}
an unbiased, low-variance baseline requiring no critic.

\paragraph{LOOP.} LOOP \citep{loop2025} generalises RLOO by applying the leave-one-out
advantage \emph{within} the PPO clipped objective, allowing the policy to drift off-policy
within a trust region and thereby reuse rollouts across $N_{\text{epoch}}$ inner epochs per
batch:
\begin{equation}
\mathcal{L}^{\text{LOOP}}(\theta)=
\hat\E_{x}\,\hat\E_{i\in[K]}\,\hat\E_{t\in y_i}
\Big[\min\big(\rho_{i,t}(\theta)\hat A_i,\;
\mathrm{clip}(\rho_{i,t}(\theta),1-\epsilon,1+\epsilon)\hat A_i\big)\Big],
\end{equation}
with $\rho_{i,t}(\theta)=\pi_\theta(a_{i,t}\mid s_{i,t})/\pi_{\theta_{\text{old}}}(a_{i,t}\mid s_{i,t})$
and the trajectory-level advantage $\hat A_i$ broadcast to every token. Following
\citet{loop2025} we drop transitions with $|\hat A_i|<\delta$ (advantage filtering).
When run purely on-policy---a single gradient update per rollout batch
($N_{\text{epoch}}{=}1$, no mini-batching)---the clipped objective reduces to REINFORCE and
LOOP recovers RLOO exactly \citep{loop2025}. This is the regime we train in: compute limits
preclude the off-policy rollout reuse that distinguishes the stronger token-level LOOP
variants, so our specialists are effectively RLOO. The reward $R\in[0,1]$ is the fraction of
the task's AppWorld unit tests passed (\S\ref{sec:setup-bench}); the binary task-goal
completion (TGC) metric we evaluate on (\S\ref{sec:prelim-stats}) is the special case in
which \emph{all} unit tests pass.

\subsection{Model merging: TIES and RAM}
\label{sec:prelim-merge}
Given $T$ specialists with task vectors $\tv_1,\dots,\tv_T$ (each $\tv_t=\theta_t-\base$),
a merge produces $\theta_{\text{merge}}=\base+\tv_{\text{merge}}$.

\paragraph{TIES.} TIES-merging \citep{yadav2023ties} reduces interference in three steps:
(i) \emph{trim} each $\tv_t$ to its top-$k\%$ magnitude coordinates; (ii) \emph{elect} a
per-coordinate sign by signed-magnitude vote
$\gamma_i=\operatorname{sign}\!\big(\sum_t \hat\tv_{t,i}\big)$, so the direction carrying the
greater total magnitude wins the coordinate; (iii) \emph{disjoint-average} only the entries
whose sign agrees with $\gamma_i$, discarding the rest,
\begin{equation}
(\tv_{\text{merge}})_i=\frac{1}{|\mathcal{A}_i|}\sum_{t\in\mathcal{A}_i}\hat\tv_{t,i},
\qquad \mathcal{A}_i=\{t:\operatorname{sign}(\hat\tv_{t,i})=\gamma_i\},
\end{equation}
scaled as $\base+\lambda\,\tv_{\text{merge}}$. For $T{=}2$ this reduces to a simple rule at
each surviving coordinate: where the two specialists agree in sign, average them; where they
conflict, the signed-magnitude vote selects the larger-magnitude entry and discards the
other. With sign agreement at chance (\S\ref{sec:results-geom}), roughly half of the
co-surviving coordinates fall in this conflict case, where the merge reduces to an arbitrary
magnitude-based selection between two near-orthogonal directions.

\paragraph{RAM.} RAM \citep{ram2026} partitions coordinates by \emph{support
co-occurrence}. With threshold $\varepsilon$, define the support mask
$m_{t,i}=\indic[\,|\tv_{t,i}|>\varepsilon\,]$ and the overlap count
$c_i=\sum_{t}m_{t,i}$. Coordinates split into \emph{shared} $\mathcal{S}=\{i:c_i\ge2\}$
and \emph{unique} $\mathcal{U}=\{i:c_i=1\}$ (coordinates with $c_i=0$ are left at base):
\begin{equation}
(\tv_{\text{merge}})_i=
\begin{cases}
\dfrac{1}{c_i}\displaystyle\sum_{t:\,m_{t,i}=1}\tv_{t,i}, & i\in\mathcal{S}\quad(\text{average shared}),\\[1.2em]
\lambda_{t(i)}\,\tv_{t(i),i}, & i\in\mathcal{U}\quad(\text{amplify unique}),
\end{cases}
\end{equation}
where $t(i)$ is the unique task touching $i$, and the amplification
$\lambda_t=1+r_{\text{\tiny RAM}}\cdot\mathrm{clip}(\rho_t,0,\alpha_{\text{\tiny clip}})$
scales with the task's shared-to-unique support ratio
$\rho_t=|\mathcal{S}_t|/|\mathcal{U}_t|$ (their Overlap--Unique Ratio); we write
$r_{\text{\tiny RAM}},\alpha_{\text{\tiny clip}}$ for RAM's rescale strength and clip bound
to avoid collision with the LoRA rank $r$ and scale $\alpha$ (\S\ref{sec:prelim-lora}). RAM+
is the rescaled variant with $r_{\text{\tiny RAM}}{=}0.1,\ \alpha_{\text{\tiny clip}}{=}2.0$
(RAM is the special case $r_{\text{\tiny RAM}}{=}0$); at these settings the amplification
saturates at $1+r_{\text{\tiny RAM}}\alpha_{\text{\tiny clip}}=1.2$. \emph{The partition
depends only on $|\tv_{t,i}|>\varepsilon$ --- the support --- and is blind to the sign or
direction of $\tv_{t,i}$.} This independence is central to \S\ref{sec:discussion}.

A caveat specific to our parameterisation: RAM is motivated by \emph{full-model} RL agents,
whose task vectors are sparse and heterogeneous in support (\citet{ram2026} report densities
from $3.2\%$ to $54.3\%$ across independently trained coding, tool-use, and memory agents),
so the unique set $\mathcal{U}$ is large and carries task-specific signal that averaging
would dilute. Our specialists are LoRA adapters: each per-module block is
$\tfrac{\alpha}{r}B_iA_i$, a product of dense factors, so every specialist modifies
essentially the same coordinates and support co-occurrence is near-total by construction
(\S\ref{sec:prelim-lora}). The unique set is correspondingly small and $\rho_t$ large; we
quantify this in \S\ref{sec:discussion} and show that RAM's rescale branch has little to act
on, so the method reduces toward the shared-set averaging it shares with TIES.

\subsection{Statistical and geometric methodology}
\label{sec:prelim-stats}
Because our effects are small on $168$ tasks, we adopt paired, distribution-free tests
and calibrate the geometric measurements.

\paragraph{Metrics.} Following AppWorld/LOOP, the headline metric is \emph{task-goal
completion} (TGC): the fraction of tasks whose unit tests \emph{all} pass (equivalently
the per-task success rate). \emph{Scenario-goal completion} (SGC) is the fraction of
scenarios in which \emph{every} task variant succeeds. Both are strict, binary,
per-task/per-scenario outcomes.

\paragraph{Paired significance (McNemar).} Two models are evaluated on the identical task
set, giving paired binary outcomes. For a pair, let $b$ be the number of tasks the first
solves and the second does not, and $c$ the reverse. Under $H_0$ (equal accuracy), the
discordant pairs satisfy $b\sim\mathrm{Binomial}(b+c,\tfrac12)$, and we report the
two-sided exact McNemar $p$-value \citep{mcnemar1947}. For contrasts against the base
model (for which we have only the official aggregate, not per-task outcomes) we fall back
to an unpaired two-sided Fisher exact test and flag it as such.

\paragraph{Bootstrap CIs.} We report $95\%$ confidence intervals for TGC and for paired
$\Delta$TGC by resampling tasks with replacement ($20{,}000$ resamples, fixed seed); a
$\Delta$TGC interval that includes $0$ indicates the contrast is within noise.

\paragraph{Task-vector geometry.} We summarise the relationship between two specialists'
task vectors by (i) the global cosine
$\cos(\tv_1,\tv_2)=\langle\tv_1,\tv_2\rangle/(\|\tv_1\|\,\|\tv_2\|)$ over all merged
coordinates, (ii) the \emph{sign agreement} on the shared support
$\{i:|\tv_{1,i}|,|\tv_{2,i}|>\varepsilon\}$, and (iii) the \emph{support overlap}
$|\{i:|\tv_{1,i}|,|\tv_{2,i}|>\varepsilon\}|$ relative to each support. In LoRA space we
compute these per module from the factors via the identity
\begin{equation}
\langle B_1A_1,\,B_2A_2\rangle_F=\big\langle B_1^{\!\top}B_2,\;A_1A_2^{\!\top}\big\rangle_F,
\label{eq:fro}
\end{equation}
which reduces all inner products to $r\times r$ contractions.

\paragraph{Floor/ceiling calibration.} A skeptic may object that low-rank adapters with
different random initialisations are near-orthogonal \emph{by construction}, so a small
cosine measures LoRA rather than learning. We bracket the measurement's dynamic range.
The \emph{floor} (null) draws two independent random rank-$r$ adapters matched to the
trained shapes and computes their cosine; we also report a \emph{spectrum-matched} null
that carries each trained task vector's singular-value profile
($\sigma=\sqrt{\mathrm{eig}((AA^{\!\top})(B^{\!\top}B))}$) along random orthonormal
directions. The \emph{ceiling} (positive control) computes the cosine between the
\emph{same} run's task vector at two training checkpoints, showing the metric can register
high alignment. A measured value well above the floor and well below the ceiling is real
signal, not a parameterisation artifact.

\section{Related work}
\label{sec:related}

\paragraph{Task arithmetic and interference.} Task vectors and their additive
composition originate with \citet{ilharco2023taskarithmetic}, who observe that
low inter-task cosine similarity is associated with reduced interference---the
hypothesis our measurements probe directly; we refer to \citet{yang2024merging}
for a comprehensive survey of the merging literature. Because a naive sum
$\sum_t \tv_t$ underperforms, a family of methods asks \emph{what} about the sum is
harmful and corrects it. \emph{Coordinate-wise} methods act in raw parameter space:
TIES \citep{yadav2023ties} trims low-magnitude entries and resolves per-coordinate
sign conflicts; DARE \citep{yu2024dare} randomly drops and rescales; Consensus/TALL
merging \citep{wang2024localizing} keeps coordinates deemed important by multiple
tasks. Weight-averaging baselines (Model Soups, \citealp{wortsman2022soups}; Fisher
and RegMean, \citealp{matena2022fisher,jin2023regmean}) average without an explicit
interference model. A distinct \emph{geometric} family argues that interference is
directional and operates on the singular structure of the per-layer task
\emph{matrices}: TSV \citep{gargiulo2025tsv} defines interference through cross-task
singular-vector overlap and removes it by orthogonalisation; AWD
\citep{xiong2024awd} subtracts a shared ``redundant'' vector so the remaining task
vectors are more orthogonal, deriving that orthogonality is sufficient for a
vanishing merging gap; Iso-C/Iso-CTS \citep{isocts2025} flatten the merged spectrum
and add task-specific subspaces orthogonal to a common one. Our results speak to
both families: within-benchmark RL specialists are already near-orthogonal
(\S\ref{sec:results-geom}), so the interference these methods target is small here,
and coordinate-wise merges collapse toward averaging (\S\ref{sec:discussion}).

\paragraph{Merging reinforced and agentic models.} Extending merging to RL-trained
agents, RAM \citep{ram2026} partitions coordinates by \emph{support co-occurrence}
and amplifies task-unique coordinates, motivated by the observation that RL task
vectors are sparse and heterogeneous across independently produced agents. RAM is
distinctive in that its partition is a function of support alone and carries no
directional information---in contrast to TIES (per-coordinate sign) and the
geometric family (singular-vector alignment). We study the regime this distinction
matters in: when specialists share support but not direction (\S\ref{sec:results-geom}),
support co-occurrence ceases to be a proxy for shared learning, and RAM's shared/unique
partition has little to act on. We note that RAM's motivating heterogeneity was measured
across agents trained by different groups with different algorithms, data, and budgets;
our within-benchmark, single-recipe setting isolates the effect of the \emph{task split}
alone, and finds dense, co-located support---a direct consequence of the LoRA
parameterisation (\S\ref{sec:prelim-lora}) rather than of what was learned.

\paragraph{A caution on cosine.} Our central diagnostic is a global cosine, and the
merging literature does not speak with one voice on it. AWD \citep{xiong2024awd}
provides direct causal evidence that global cosine drives merge quality---amplifying
inter-task cosine while holding individual task performance fixed collapses the merged
model---which supports reading our near-zero cosine as a benign regime. Iso-CTS
\citep{isocts2025}, however, reports that cosine correlates only weakly with merge
gain and that a small global cosine can coexist with substantial overlap of the
\emph{top singular subspaces}, where functional interference concentrates. Our
floor/ceiling calibration (\S\ref{sec:results-calib}) addresses the objection that a
small cosine is a low-rank artifact, but not the distinct objection that global cosine
understates subspace overlap; we flag the latter as a limitation
(\S\ref{sec:discussion}) and regard subspace-level measurement (SAR, STI) on RL
specialists as the natural follow-up.

\paragraph{Training for mergeability, and decentralised training.} A separate line
modifies training so that the resulting models merge better---\emph{pre-hoc} merging.
Orthogonal Adaptation \citep{po2024orthogonal} enforces near-orthogonal updates during
fine-tuning; tangent-space fine-tuning \citep{ortiz2023tangent} and partial
linearisation \citep{tang2024llora} improve weight disentanglement; ATM
\citep{zhou2024atm} alternates tuning and merging. These change the loss, optimiser,
or parameterisation. Our future-work objective (\S\ref{sec:future}) instead asks only
\emph{when} to stop, and is thus deployable across independent workers---connecting to
communication-efficient parallel training, where independently trained experts are
combined post hoc: Branch-Train-Merge \citep{li2022btm} and Branch-Train-MiX
\citep{sukhbaatar2024btx} for expert LMs, and federated / low-communication schemes
such as FedAvg \citep{mcmahan2017fedavg} and DiLoCo \citep{douillard2023diloco} for
distributed optimisation. These shard \emph{data}; the agentic-RL setting instead
shards \emph{environments}, which is the constraint that makes joint on-policy training
costly and post-hoc merging attractive. We do not pursue this direction empirically
here (\S\ref{sec:future}), but it motivates why the merge-vs-joint comparison is worth
establishing.

\section{Experimental setup}
\label{sec:setup}

\subsection{Benchmark and metrics}
\label{sec:setup-bench}
AppWorld \citep{trivedi2024appworld} is an execution-based benchmark of interactive
digital tasks spanning multiple simulated apps; each task is graded by hidden unit tests.
We evaluate on the held-out \texttt{test\_normal} split of $168$ tasks, composed of
$57$ difficulty-1, $48$ difficulty-2, and $63$ difficulty-3 tasks. Difficulty-3 tasks are
essentially unsolved by all models here ($\le1.6\%$ TGC) and dilute the aggregate; the
per-difficulty view on d1/d2 is the informative one. We note that the \texttt{test\_challenge} benchmark exists in AppWorld, however, smaller scale models have too little signal to effectively optimize on, so this split is left out of the analysis. Individual tasks are inherently
multi-app, so specialisation is by \emph{difficulty}, not by app. The reward for RL is the
task outcome; TGC/SGC are as defined in \S\ref{sec:prelim-stats}.

\subsection{Models and training}
\label{sec:setup-train}
All models start from Qwen3-8B and are trained with LOOP in its on-policy regime
(RLOO; \S\ref{sec:prelim-rl}) using LoRA rank $r{=}16$ adapters on the attention and
MLP projection matrices
($\{q,k,v,o,\text{gate},\text{up},\text{down}\}\_\text{proj}$; $252$ adapted modules).
Key hyperparameters follow \citet{loop2025}: $K{=}6$ rollouts per task, sampling
temperature $1.0$, learning rate $5\times10^{-5}$, one update epoch per batch
($N_{\text{epoch}}{=}1$), advantage filter $\delta{=}0.01$, and a per-episode interaction
budget of $40$. We train three models under an identical recipe:
\begin{itemize}
\item \textbf{Difficulty-1 specialist} --- trained only on difficulty-1 tasks.
\item \textbf{Difficulty-2 specialist} --- trained only on difficulty-2 tasks.
\item \textbf{Joint} --- trained on the union of difficulty-1 and difficulty-2 tasks.
\end{itemize}
Each is run for $\sim\!10$ RL iterations; peak checkpoints are selected on a held-out
criterion (specialists: iteration $5$; joint: iteration $9$). Training uses $6\times16$\,GB
GPUs with a balanced device map and gradient checkpointing.

\subsection{Merging}
\label{sec:setup-merge}
We merge the two specialists' peak checkpoints (initially using peak as defined as top performance on the train set) with (i) TIES \citep{yadav2023ties} and
(ii) RAM+ \citep{ram2026}, both applied to the
full merged task vectors. Merging is performed in a separate environment
(\texttt{transformers} 4.51--4.57 for Qwen3 compatibility).

\subsection{Evaluation protocol (reproducibility)}
\label{sec:setup-eval}
Evaluation is deterministic by construction: \emph{greedy} decoding (temperature $0$),
fixed random seed $100$, and a fixed AppWorld seed, served by vLLM (tensor-parallel $2$,
context length $32{,}768$). Each task is evaluated in an isolated AppWorld instance; TGC,
SGC, and per-difficulty breakdowns are computed post hoc from per-task outcome logs. Code
for training, merging, evaluation, all statistics, and the geometry calibration is released. The only non-determinism we do not eliminate is
vLLM's scheduling/floating-point reduction order; a single confirmatory re-run would
bound it. 

\section{Results}
\label{sec:results}

\subsection{Performance: merges, specialists, and joint RL all tie}
\label{sec:results-perf}
\begin{figure}[t]
\centering
\includegraphics[width=0.72\linewidth]{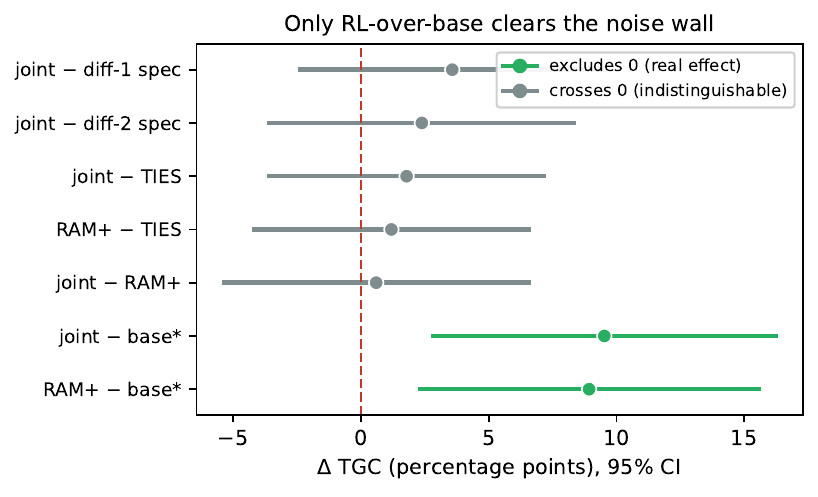}
\caption{Paired $\Delta$TGC with $95\%$ bootstrap CIs. Every merge/joint contrast crosses
$0$ (indistinguishable); only RL-over-base excludes $0$. The interesting comparisons live
below the noise wall of this benchmark.}
\label{fig:forest}
\end{figure}

Table~\ref{tab:main} reports per-difficulty and aggregate TGC (with $95\%$ bootstrap CIs)
and SGC. RL lifts aggregate TGC from the $6.6\%$ base to a $12$--$16\%$ band, but within
that band the six models overlap heavily. Table~\ref{tab:sig} shows no paired contrast
among the RL models is significant (McNemar $p\ge0.56$), while every RL model beats the
base (unpaired Fisher $p\le0.05$). Figure~\ref{fig:forest} makes the picture visual: every
merge/joint $\Delta$TGC CI crosses $0$; only RL-over-base clears it.

Two consequences follow. First, \emph{behaviour-aware merging is not distinguishable from
naive merging here}: RAM+ vs.\ TIES is a net one-task difference ($p{=}0.83$). Second, and
more useful, \emph{merging is indistinguishable from joint multi-task RL}
(joint vs.\ RAM+ $p{=}1.00$; joint vs.\ TIES $p{=}0.66$) --- the comparison the
cross-domain RL-merging setting cannot make, since it lacks joint-training data. Merging
recovers what joint RL achieves on the same data, to within the resolution of this
benchmark.

\begin{table}[t]
\centering
\caption{Held-out \texttt{test\_normal} TGC (\%) by difficulty, aggregate TGC with $95\%$
bootstrap CI, and SGC (\%). Base is the official external report (no per-task file). All
RL variants fall in one overlapping band.}
\label{tab:main}
\small
\begin{tabular}{lccccc}
\toprule
Model & d1 & d2 & d3 & Agg.\ TGC [95\% CI] & SGC \\
\midrule
Base                 & 15.8 & 2.1  & 1.6 & 6.6\ \ (external) & 0.0 \\
Difficulty-1 spec.\  & 31.6 & 4.2  & 1.6 & 12.5\ \ [7.7,\,17.9] & 5.4 \\
Difficulty-2 spec.\  & 29.8 & 10.4 & 1.6 & 13.7\ \ [8.9,\,19.0] & 3.6 \\
Joint                & 36.8 & 10.4 & 1.6 & 16.1\ \ [10.7,\,21.4] & 3.6 \\
TIES merge           & 35.1 & 8.3  & 0.0 & 14.3\ \ [9.5,\,19.6] & 5.4 \\
RAM+ merge           & 36.8 & 10.4 & 0.0 & 15.5\ \ [10.1,\,20.8] & 7.1 \\
\bottomrule
\end{tabular}
\end{table}

\begin{table}[t]
\centering
\caption{Pairwise significance on TGC. \emph{Top:} paired McNemar (exact) among RL
models; the two counts are \emph{discordant} tasks (A-solves-not-B / B-solves-not-A).
\emph{Bottom:} base has no per-task file (external report), so these are unpaired Fisher
exact tests and the two counts are \emph{marginal} successes (A / base, out of $168$) ---
not discordant pairs. No RL--RL contrast is significant; every RL model beats base.}
\label{tab:sig}
\small
\begin{tabular}{llccc}
\toprule
Contrast (A vs.\ B) & test & count A & count B & $p$ \\
\midrule
Joint vs.\ RAM+            & McNemar & 14 & 13 & 1.000 \\
RAM+ vs.\ TIES             & McNemar & 12 & 10 & 0.832 \\
Joint vs.\ TIES            & McNemar & 12 & \ 9 & 0.664 \\
Difficulty-1 vs.\ Diff-2   & McNemar & 11 & 13 & 0.839 \\
Joint vs.\ Difficulty-2    & McNemar & 15 & 11 & 0.557 \\
\midrule
RAM+ vs.\ Base             & Fisher  & 26 & 11 & 0.014 \\
Joint vs.\ Base            & Fisher  & 27 & 11 & 0.009 \\
Difficulty-2 vs.\ Base     & Fisher  & 23 & 11 & 0.045 \\
\bottomrule
\end{tabular}
\end{table}

\subsection{Task-vector geometry and its development}
\label{sec:results-geom}
\begin{figure}[t]
\centering
\includegraphics[width=0.62\linewidth]{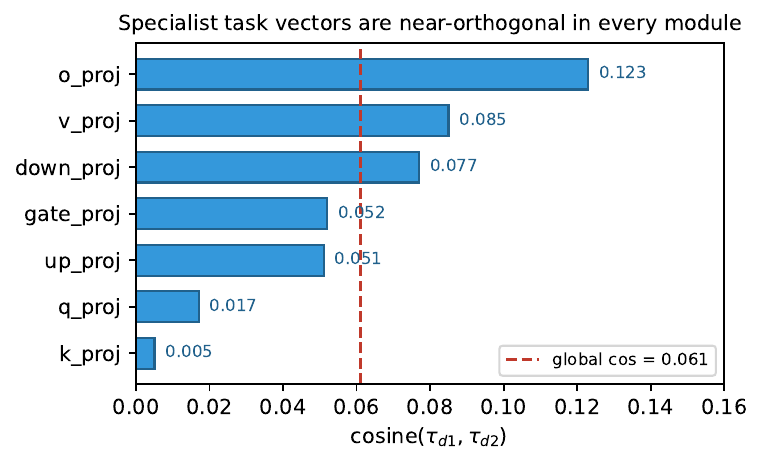}
\caption{Per-module cosine between the two specialists' task vectors. Near-orthogonality
is uniform across module types (global $0.061$; maximum $0.12$).}
\label{fig:module}
\end{figure}

Why is there so little for behaviour-aware merging to gain? We examine the specialists'
task vectors. Although their supports overlap by $\sim\!65\%$ (a consequence of adapting
identical LoRA target matrices, \S\ref{sec:prelim-lora}), their \emph{directions} are
nearly independent: the global cosine is $0.06$--$0.10$, and sign agreement on the shared
support is $51.9\%$ --- chance. Figure~\ref{fig:module} shows this holds in every module
type (maximum $0.12$ in $o\_\text{proj}$). Support overlap and directional alignment are
thus effectively decoupled.

\begin{figure}[t]
\centering
\includegraphics[width=0.62\linewidth]{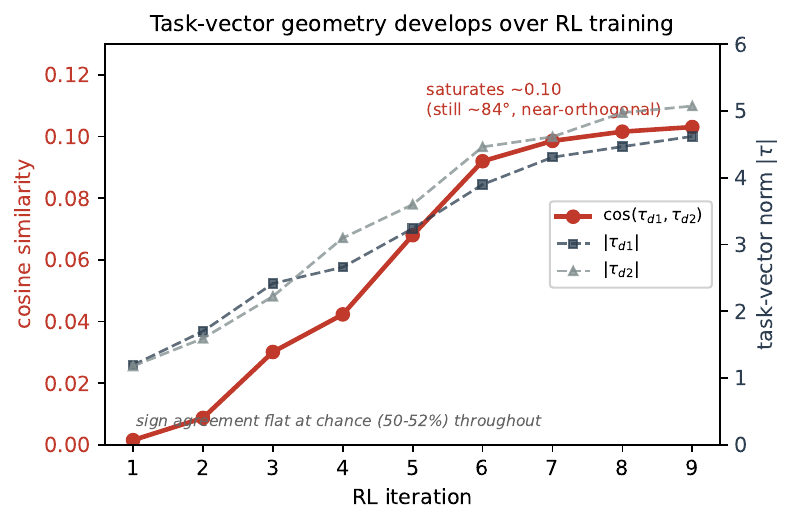}
\caption{Task-vector geometry develops over RL training. $\cos(\tv_{d1},\tv_{d2})$ (red)
rises from $\approx0$ to $\approx0.10$ and saturates; $\|\tv\|$ (grey, right axis) grows
$\sim\!4\times$; sign agreement stays at chance throughout.}
\label{fig:traj}
\end{figure}

This small shared component is not present from the start but \emph{develops}:
Figure~\ref{fig:traj} tracks $\cos(\tv_{d1},\tv_{d2})$ across training. It rises
monotonically from $0.002$ (iteration $1$) to $0.103$ (iteration $9$) and saturates near
$0.10$ ($\approx84^\circ$, still dominantly orthogonal), while $\|\tv\|$ grows roughly
fourfold and sign agreement stays flat at chance. The shape mirrors the known growth of
inter-task \emph{gradient} alignment in multi-task learning, here shown in weight space
for RL specialists; we do not claim the phenomenon is novel, but its measurement fixes the
regime in which support-based merging operates.

\subsection{The measurement is real, not a low-rank artifact}
\label{sec:results-calib}
\begin{figure}[t]
\centering
\includegraphics[width=0.72\linewidth]{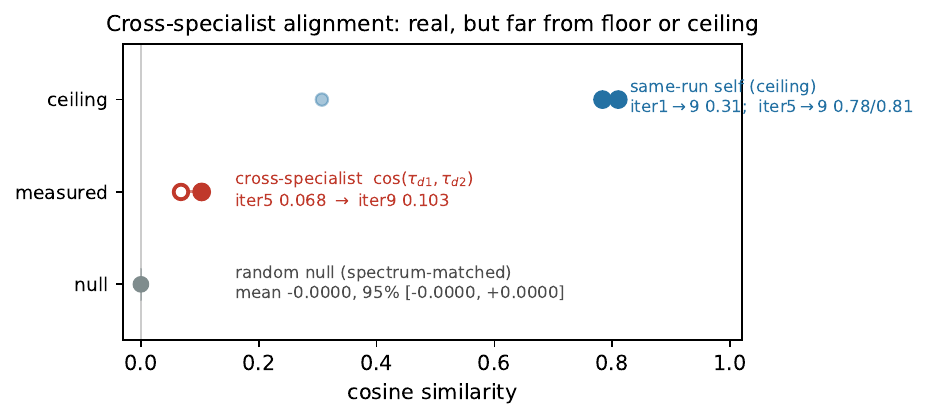}
\caption{Floor/ceiling calibration. Random and spectrum-matched rank-$16$ nulls give
$|\cos|<10^{-4}$ (floor); same-run self-alignment reaches $\approx0.80$ (ceiling); the
cross-specialist measurement ($0.068\to0.103$) lies well between. Near-orthogonality
reflects learning, not the low-rank parameterisation.}
\label{fig:calib}
\end{figure}

To rule out that $0.06$--$0.10$ is imposed by the rank-$16$ parameterisation, we calibrate
against a floor and a ceiling (Figure~\ref{fig:calib}). Two independent random rank-$16$
adapters, matched to the trained shapes, give $|\cos|<10^{-4}$; matching the trained
singular-value spectrum does not change this (std $\approx2\times10^{-5}$), because in this
ambient dimension random directions concentrate near-orthogonally regardless of spectrum.
The same run's task vector at two checkpoints, by contrast, reaches $\cos\approx0.78$--$0.81$,
so the metric is far from floor-pinned. The cross-specialist value therefore sits
two-to-three orders of magnitude above the floor and $\sim\!8\times$ below the ceiling: the
specialists are genuinely, but not trivially, near-orthogonal. The developing trajectory
of \S\ref{sec:results-geom} corroborates this --- a construction artifact would sit at the
floor for all iterations rather than climbing with training. \emph{We deliberately avoid
reporting standard-score (``$\sigma$'') distances from the null: because the null variance
is so small, such scores are inflated and uninformative.}

\subsection{A finer metric, and the over-readings it invites}
\label{sec:results-metric}
The TGC null (\S\ref{sec:results-perf}) is partly a resolution limit: binary all-or-nothing
outcomes on $168$ tasks, tested with McNemar (which uses only the $\sim\!20$--$27$
discordant pairs), have little power. AppWorld also exposes a continuous per-task score
(fraction of unit tests passed); a paired test on all $168$ continuous differences is far
more sensitive. We report it not because it overturns the null, but because it is
instructive about \emph{how} such a finding can mislead --- and we walk through three
over-readings we ourselves entertained and discarded.

Table~\ref{tab:metric} shows the continuous score does resolve differences TGC cannot:
joint and both specialists significantly outscore the merges (Wilcoxon $p\le0.016$),
while behaviour-aware still ties naive (RAM+ vs.\ TIES $p{=}0.57$) and joint ties its
diff-2 component ($p{=}0.71$). Taken alone this reads as ``merging costs partial
progress.'' Three checks show why that reading is unsafe.

\emph{(1) It reverses across metrics.} On TGC the merges \emph{beat} diff-2 alone
($14.3,15.5$ vs.\ $13.7$); on the continuous score they fall \emph{below} it
($0.45$ vs.\ $0.52$). Merging recombines discrete completions but scores lower on diffuse
partial progress. A direction that flips with the metric cannot support a capability
claim.

\emph{(2) It is not emergent cross-task capability.} A natural story is that joint
co-training discovers cross-task capability absent from either specialist. It does not:
$\mathrm{mean}(\text{joint})=0.539$ is \emph{below} the mean per-task best specialist
$\mathrm{mean}(\max(\tau_{d1},\tau_{d2}))=0.581$, and joint lies within
$[\min,\max]$ of the two specialists on $120/168$ tasks. Joint approximates
best-of-specialists; it does not exceed it. And a \emph{single} specialist (diff-2) also
beats the merges, so the gap is a property of \emph{merging}, not of joint co-training.

\emph{(3) The merges track the per-task average.} Both merges are closer to the per-task
\emph{average} of the specialists' scores than to the per-task max
(Table~\ref{tab:metric}, bottom), i.e.\ merging $\approx$ averaging competence, which is
mechanically below best-of. This is the same near-orthogonal-averaging picture as
\S\ref{sec:discussion}, now visible in outputs.

The honest summary: the continuous metric reveals that merging trades a little diffuse
partial-progress behaviour for equal completions, consistent with averaging
near-orthogonal task vectors --- a mechanism observation, not evidence that merging fails
as a joint-training substitute. On the metric the benchmark actually reports (TGC),
merging matches joint.

\begin{table}[t]
\centering
\caption{Continuous-metric analysis. \emph{Top:} per-model TGC (\%) vs.\ mean partial-credit
score --- note the reversal for diff-2 vs.\ the merges. \emph{Middle:} paired tests for key
contrasts (McNemar on TGC vs.\ Wilcoxon signed-rank on the continuous score); the
continuous metric resolves what TGC cannot. \emph{Bottom:} the merges track the per-task
average of the specialists, not the max (mean absolute deviation, lower = closer);
joint sits below the per-task max, so there is no emergence.}
\label{tab:metric}
\small
\begin{tabular}{lcc}
\toprule
Model & TGC (\%) & mean score \\
\midrule
Difficulty-1 spec. & 12.5 & 0.379 \\
Difficulty-2 spec. & 13.7 & 0.523 \\
Joint              & 16.1 & 0.539 \\
TIES merge         & 14.3 & 0.446 \\
RAM+ merge         & 15.5 & 0.459 \\
\midrule
\emph{paired contrast} & McNemar $p$ & Wilcoxon $p$ \\
 & (TGC) & (score) \\
Joint vs.\ RAM+       & 1.000 & 0.005 \\
Joint vs.\ TIES       & 0.664 & 0.001 \\
RAM+ vs.\ TIES        & 0.832 & 0.567 \\
Diff-2 vs.\ RAM+      & 0.701 & 0.016 \\
Diff-2 vs.\ TIES      & 1.000 & 0.005 \\
Joint vs.\ Diff-2     & 0.557 & 0.707 \\
\midrule
\emph{emergence / averaging} & \multicolumn{2}{c}{value} \\
mean(joint) vs.\ per-task max & \multicolumn{2}{c}{$0.539 < 0.581$ (no emergence)} \\
TIES $|{\cdot}\!-\!\mathrm{avg}|$ vs.\ $|{\cdot}\!-\!\max|$ & \multicolumn{2}{c}{$0.226 < 0.264$ (tracks avg)} \\
RAM+ $|{\cdot}\!-\!\mathrm{avg}|$ vs.\ $|{\cdot}\!-\!\max|$ & \multicolumn{2}{c}{$0.208 < 0.241$ (tracks avg)} \\
\bottomrule
\end{tabular}
\end{table}

\section{Discussion}
\label{sec:discussion}

\paragraph{Why RAM has little to disentangle here.} RAM partitions coordinates by support
co-occurrence and \emph{averages} the shared set (\S\ref{sec:prelim-merge}). With
$\sim\!65\%$ support overlap but near-zero directional alignment, most co-located updates
are directionally independent, yet RAM labels them ``shared'' and averages them --- halving
signal that is effectively task-specific. Its unique set is correspondingly
under-populated relative to the true (directional) geometry, and on these models the
count-based amplification is mild. The net effect is that RAM behaves close to uniform
averaging, which is consistent with RAM+, TIES, and the naive/joint outcomes all landing
in one statistical band (\S\ref{sec:results-perf}). We stress this is an argument about
\emph{this regime} --- support-aware merging may well help where supports genuinely encode
shared vs.\ private directions; our point is that support co-occurrence is not a reliable
proxy for direction, and here the two come apart.

\paragraph{Merging vs.\ joint RL.} The one comparison our within-benchmark design enables
and the cross-domain setting cannot --- merge vs.\ joint on the same data --- comes out a
tie. For practitioners who \emph{can} pool data, merging buys nothing over joint RL at this
scale; for those who cannot, merging at least does not obviously sacrifice what joint RL
would have delivered.

\paragraph{Peak-checkpoint selection.} Per arm we merged and evaluated the checkpoint of
maximum mean \emph{training} reward (diff-1 iter5, diff-2 iter5, joint iter9), \emph{not}
a checkpoint chosen on \texttt{test\_normal}; RL was unstable past the peak (reward
collapsed after iter5, $0.569\!\to\!0.387$ and $0.464\!\to\!0.40$), so later checkpoints
were not used. This is mild optimistic selection on in-distribution reward, not test
leakage --- the reported test numbers are held out from selection.

\paragraph{Limitations.} (i) Single seed per model: our claims are indistinguishability at
the resolution of $168$ tasks, not proven equivalence. This matters most for the
continuous-metric result of \S\ref{sec:results-metric}: before that signal could support
any capability claim it would need to clear joint-vs-joint seed variance (a conservative
free proxy is joint's neighbouring checkpoints evaluated on the continuous metric). We
therefore report it as metric-sensitivity, not capability. (ii) Scale: $8$B parameters,
$\sim\!10$ RL iterations; larger effects may emerge with more training or larger models.
(iii) One specialist pair on one benchmark: the geometry is a single, if clean and
calibrated, observation; whether near-orthogonality persists across other splits or
app-level specialisations is open. (iv) The partial-credit metric is softer than TGC and
we lean on it only for the negative/mechanistic points of \S\ref{sec:results-metric}.

\section{Future work: merge-aware checkpoint early stopping}
\label{sec:future}
Every merging method \citep{yang2024merging} takes task vectors as fixed input: one point per task, harvested
from a final checkpoint. But an RL run produces a \emph{trajectory} $\tv_t(s)$, and which
point one hands to the merge operator is a free parameter. Two quantities grow together
over training (\S\ref{sec:results-geom}): capability $\|\tv_t(s)\|$ (reward rises to a
peak) and interference $\cos(\tv_{d1}(s_1),\tv_{d2}(s_2))$. Merge quality is plausibly a
difference of the two, so one might stop each specialist early --- trading a little
capability for lower interference --- via an objective
\begin{equation}
J(s_1,s_2)=\underbrace{P_1(s_1)+P_2(s_2)}_{\text{capability (training reward)}}
-\;\gamma\underbrace{\,I(s_1,s_2)}_{\text{interference}},
\label{eq:jobj}
\end{equation}
with $P_t$ the mean training reward already logged and $I$ computed from the LoRA factors
(no evaluation needed). Selection need not be diagonal ($s_1\neq s_2$ is allowed), and in
a decentralised setting each worker could stop on its own signal without synchronisation. Modifying training to improve mergeability has precedent \citep{po2024orthogonal, zhou2024atm} (\S\ref{sec:related}); our variant intervenes only on the \emph{stopping time}, changing neither the loss nor the parameterisation, and is therefore deployable across independent workers.

\begin{figure}[t]
\centering
\includegraphics[width=0.56\linewidth]{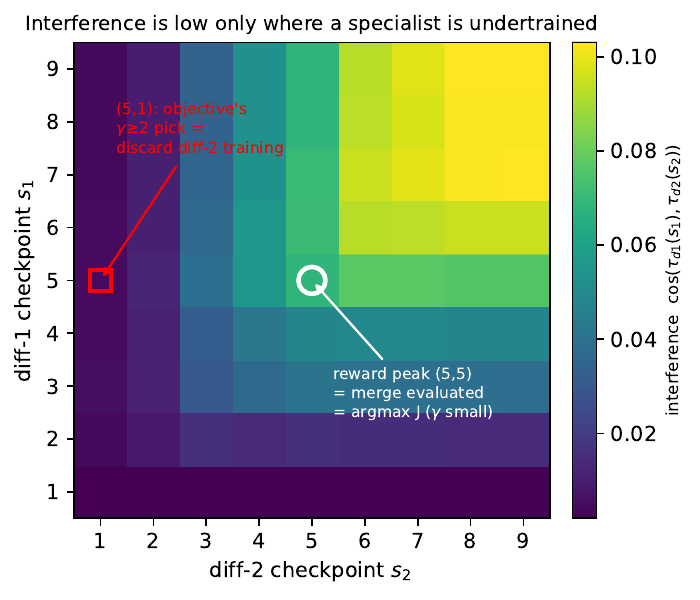}
\caption{Interference grid $I(s_1,s_2)=\cos(\tv_{d1}(s_1),\tv_{d2}(s_2))$ over training
checkpoints. Low interference coincides with an undertrained specialist (dark corners);
it rises toward the capability peak $(5,5)$, the merge actually evaluated and the argmax of
Eq.~\ref{eq:jobj} for any reasonable $\gamma$. The objective's only low-interference
alternative, $(5,1)$, discards a specialist's training. Merge-aware early stopping has
nothing to select here.}
\label{fig:jgrid}
\end{figure}

\emph{In our regime this cannot help, and the reason is instructive.}
Figure~\ref{fig:jgrid} shows the full interference grid $I(s_1,s_2)$: interference is low
\emph{only} in the undertrained corner and rises with both indices toward the
capability peak $(5,5)$. Consequently the argmax of Eq.~\ref{eq:jobj} is $(5,5)$ --- the
reward-peak pair we already merged --- for all $\gamma$ below $\sim\!4$; at $\gamma=0$
(pure capability) it is $(5,5)$ by definition, and no grid cell exceeds that capability.
Any $\gamma$ large enough to move the selection lands on $(5,1)$, which merely
\emph{discards} diff-2's training (its low interference is the triviality of a tiny,
undeveloped task vector). There is no high-capability, low-interference cell to select,
because in this regime interference and capability are both driven by $\|\tv\|$ growth and
rise together. The merge already evaluated is the objective's optimum.

We conjecture the lever has leverage only where interference can be reduced
\emph{without} sacrificing capability --- larger $T$ (interference accumulates across more
specialists), cross-domain merging (larger, more-aligned task vectors), or full
fine-tuning rather than tiny LoRA deltas. Characterising that regime, and validating
$J$ against measured merge quality with a metric sensitive enough to resolve it
(\S\ref{sec:results-metric}), is the natural next step.

\section{Conclusion}
\label{sec:conclusion}
On a controlled within-benchmark AppWorld setup, merging RL specialists is statistically
indistinguishable from joint multi-task RL on the benchmark's primary metric (TGC), and
behaviour-aware merging is indistinguishable from naive merging --- all within a noise wall
that $1$--$2$ task margins cannot cross. A calibrated geometric diagnostic explains the
null: within-benchmark RL specialists remain near-orthogonal in direction (a small,
developing shared component) despite large support overlap, so support- and sign-based
merging have little private structure to protect. A finer continuous metric resolves a
merged-vs-unmerged difference, but we show it reverses across metrics, is not emergent
cross-task capability, and cannot be closed by trajectory-aware checkpoint selection ---
a set of disciplined negative results we treat as part of the contribution. We release
code and statistics to make the null, its calibration, and these checks fully
reproducible.
\bibliographystyle{tmlr} 
\bibliography{references}
\end{document}